\documentclass{article}

\PassOptionsToPackage{sort, square, numbers}{natbib}

\usepackage[final]{neurips_2023}

\usepackage[utf8]{inputenc} 
\usepackage[T1]{fontenc}    
\usepackage{hyperref}       
\usepackage{url}            
\usepackage{booktabs}       
\usepackage{amsfonts}       
\usepackage{nicefrac}       
\usepackage{microtype}      
\usepackage{xcolor}         
\usepackage[pdftex]{graphicx} 
\usepackage{notoccite}
\usepackage{lscape}

\title{Enabling Decision-Support Systems through Automated Cell Tower Detection}

\author{
  Natasha Krell$^{1*}$ Will Gleave$^{2*}$ Daniel Nakada$^{2*}$ Justin Downes$^{2}$\\
   \textbf{Amanda Willet}$^{3}$ \textbf{Matthew Baran}$^{4}$ \\
  \textsuperscript{*}Equal contribution.\\ 
  \textsuperscript{1}National Geospatial-Intelligence Agency  \textsuperscript{2}Amazon Web Services \\ \textsuperscript{3}Department of Geosciences, The Pennsylvania State University \\ \textsuperscript{4}Applied Research Laboratory, The Pennsylvania State University}

\begin{document}

\maketitle

\begin{abstract}
Cell phone coverage and high-speed service gaps persist in rural areas in sub-Saharan Africa, impacting public access to mobile-based financial, educational, and humanitarian services. Improving maps of telecommunications infrastructure can help inform strategies to eliminate gaps in mobile coverage. Deep neural networks, paired with remote sensing images, can be used for object detection of cell towers and eliminate the need for inefficient and burdensome manual mapping to find objects over large geographic regions.  In this study, we demonstrate a partially automated workflow to train an object detection model to locate cell towers using OpenStreetMap (OSM) features and high-resolution Maxar imagery. For model fine-tuning and evaluation, we curated a diverse dataset of over 6,000 unique images of cell towers in 26 countries in eastern, southern, and central Africa using automatically generated annotations from OSM points. Our model achieves an average precision at 50\% Intersection over Union (IoU) (AP@50) of 81.2 with good performance across different geographies and out-of-sample testing. Accurate localization of cell towers can yield more accurate cell coverage maps, in turn enabling improved delivery of digital services for decision-support applications.
\end{abstract}

\section{Introduction}

Improving access to mobile phones and expanding cell phone coverage serves many sustainable development goals \cite{rotondi2020leveraging}. Access to cellular networks and mobile internet is essential for many humanitarian assistance and disaster relief digital services. In rural areas in sub-Saharan Africa, large coverage gaps and a lack of high-speed services persists \cite{mehrabi2021global} and limits the diffusion of mobile-phone based services, such as climate advisories and extreme weather alerts \cite{krell2021smallholder}.

Resolving coverage gaps requires improved information on cell tower (i.e., base stations) locations, particularly in rural and remote areas. While mobile phone coverage maps, such as the Global System for Mobile Communications (GSMA), are freely available, validated cell tower locations are rarely available in the public domain. Obtaining communication tower locations from Mobile Network Operators (MNOs) or Telecommunications Companies (telecoms) is largely impossible because telecoms treat infrastructure information, specifically cell tower locations, as sensitive because it improves their market position \cite{fida2019uncovering} causing researchers to use crowdsourced and imprecise cell tower locations. Currently, the only open-source information on cell phone tower locations comes from data crowdsourced (e.g., OpenCelliD), which are often incomplete for rural and remote regions \cite{fida2019uncovering}. 


Computer vision approaches applied to high-resolution satellite imagery can detect a variety of human-made objects with high accuracy and assist in labor-intensive searches over large swaths of the Earth \cite{lam2018xview}. Deep learning models have been successfully used for a variety of tower-like objects (e.g., Yao, et al \cite{yao2017chimney}), particularly for object detection. Cell phone towers are highly variable in their physical appearance and background environment. To discover these towers across heterogeneous landscapes in a broad area search, each variation needs to be learned by the machine learning model. With the increasing availability of high-resolution satellites, cell tower detection is now possible; however, it remains challenging due to such factors as background environments, shadows, imaging angles, the lattice nature of the structures, and variations in the types of towers. Deep neural networks paired with remote sensing images can automatically learn target features and better generalize across high variance targets than other detection methods \cite{cheng2016survey, zhang2016weakly, zou2016ship}. To date, we have not discovered any published attempts to leverage machine learning to find cell towers in remotely sensed imagery.

In this study, we used a machine learning workflow for object detection of cell towers, incorporating self-supervised pretrained models and an automated bounding box generator. First, human annotators verified annotated imagery using OSM data points and automatically generated bounding boxes from the towers' center points \cite{inder2022centerpoints}. We then trained a Faster Region-based Convolutional Neural Network (Faster RCNN) architecture \cite{FasterRCNN} and tested it on electro-optical satellite imagery. Next, we trained ResNet model backbones with different weights: random initialized, ImageNet pretrained, and hierarchical pretrained \cite{DBLP:journals/corr/HeZRS15}. Our results show that cell towers can be detected accurately across a wide geographical area, which can help improve cell phone coverage maps in remote locations.  

\begin{figure}[t]
\begin{center}
   \includegraphics[width=\linewidth]{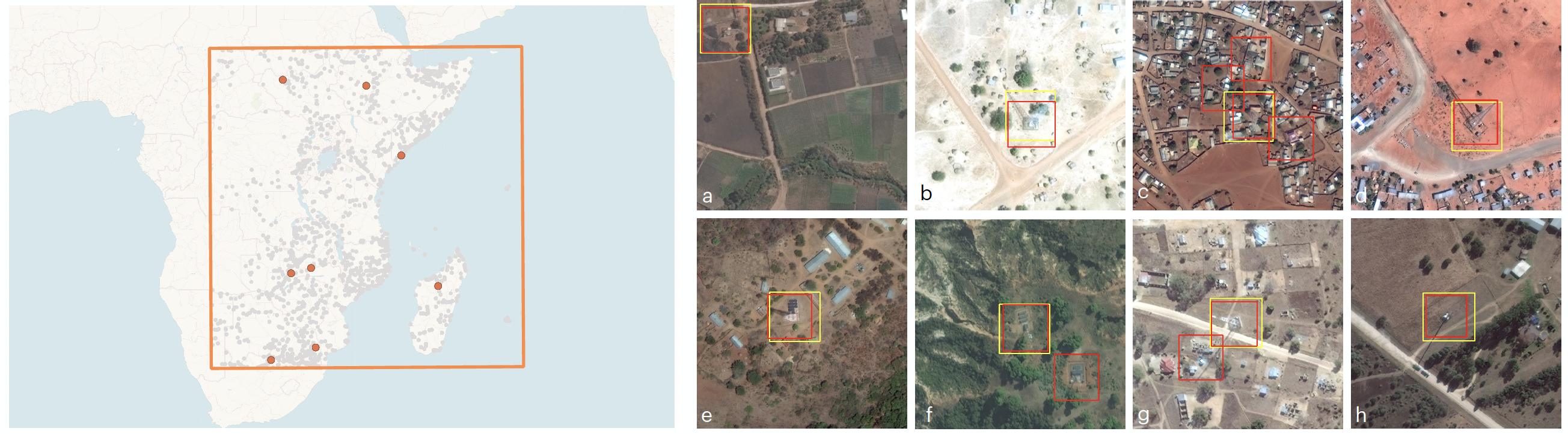}
\end{center}
   \caption{Map of study area (orange box) with cell tower locations (gray points) used in training and testing. Examples of object detection results are shown as orange points. Ground truth bounding boxes are indicated by the yellow bounding boxes. Predictions are indicated by the red bounding boxes. Labels are generated by the approach detailed in Section \ref{sec:data}. Example images are from: a) Lusaka, Zambia; b) Aweil North, South Sudan; c) Welkite, Ethiopia; d) Kismayo, Somalia; e) Nyimba, Zambia; f) Maevatanana, Madagascar; g) Seville, South Africa; and h) Kaalpan, South Africa. Images \copyright  2020-2022 Maxar.}
   \label{fig:objdet}
\end{figure}

\section{Application Context}

Mobile phone ownership is critical for improving the lives and livelihoods of those living in rural poverty in sub-Saharan Africa. A mobile phone allows access to information and opportunities, such as selling products, listening to weather forecasts, banking and saving through mobile money, generating income, and accessing health care, and educational resources. However, fully realizing the potential of these mobile-based services and maximizing their impact requires adequate cell phone coverage \cite{usaid2020}. 

Approximately 16\% of Africans live in areas without mobile network coverage with the largest gaps existing in the poorest regions \cite{gsma2022}. Telecommunications companies build telecommunications towers where customers demand a signal. In the poorest regions, mobile network coverage and usage are the most limited, for example, in central Africa where over a third of adults are not covered by a network \cite{gsma2022}. The gap in coverage is due to poor communication infrastructure, such as a lack of cell towers, fiber-optic cables, and data centers \cite{usaid2020}.

Open-source cell phone coverage maps are frequently outdated, are made at poor spatial resolutions, and can be misleading. This inaccuracy creates challenges for local development actors and humanitarian aid workers who depend on good cell phone coverage for local operations and planning. Without accurate, high-resolution cell phone coverage maps, humanitarian and development activities can be rendered ineffective and the increasing availability of mobile-phone applications may not reach end users when the cell network is not adequate.  This research is the first to create an object detector for cell towers in sub-Saharan Africa, which is an important step towards mapping telecommunications infrastructure and cell coverage gaps in remote and at-risk locations. 

\begin{table}[h]
\caption{Metrics of baseline model results. }
\label{tab:baseline}
  \centering
  \begin{tabular}{ l ccc} 
    \toprule
 Model& AP & AP@15&AP@50\\
    \midrule
 ResNet50-random initialize  & 42.1    & 78.6&  72.7 \\
 ResNet50-ImageNet pretrained &   44.4  & 85.6   & 81.2\\
 RestNet50-hierarchical pretrained & 49 & 87.5&  81.2\\
 ResNet101-ImageNet pretrained & 49.9 & 87.3& 83.7\\
    \bottomrule
  \end{tabular}
\end{table}
\section{Data}
\label{sec:data}

\paragraph{Cell Tower Labels and Imagery.} 

Deep learning models largely require high-resolution imagery for model training \cite{hamaguchi2018building}. We focused our study on cell towers in eastern and southern Africa, where gaps in mobile phone coverage exist in rural locations.  For our tower annotations, we queried OSM tags for attributes ``man\_made'' and ``communications\_tower'' or ``tower.'' We did not query over the entire continent; instead we selected an area that covers parts of eastern, southern, and central Africa, specifically from 12\textdegree N to 27\textdegree S and 20\textdegree E to 57\textdegree E. This region spans roughly a 4500 x 4000 km\textsuperscript{2} area. To focus on rural cell towers, we removed any OSM point coordinates that fell inside the Global Human Settlement layer of the Urban Centres Database (GHS-UCDB) \cite{JRC115586}. Human annotators then verified and placed OSM tower coordinates at the center of the base of the tower structure. We then downloaded WorldView-3, WorldView-2, and GeoEye-1 satellite imagery from Maxar. We queried pan-sharpened natural color images with cloud cover less than 10\%. We resampled the full scenes to 0.5 meter Ground Sample Distance (GSD) to limit variations in the scale of the target. Section A.1 outlines steps taken to generate chips and automated labels set 25 meters from towers' center points.


\section{Methodology}

\paragraph{Object Detection.}

Convolutional Neural Networks (CNNs) are commonly applied to image-related machine learning tasks, such as image classification \cite{DBLP:journals/corr/HeZRS15} and object detection \cite{ DBLP:journals/corr/RenHG015}. For our downstream object detection network, we employed Faster RCNN, a well-established architecture where the image first passes through a backbone network that produces a feature map as an input to a region proposal network. The region proposal network then generates the proposed locations of interest within the image. The backbone network is usually a dense convolutional network, such as ResNet-50 or ResNet-101. 

For our experiments, we investigated two backbones---ResNet-50 or ResNet-101---with selected weight initialization: ResNet-50 random initialized; ResNet-50 pre-trained on ImageNet classification; ResNet-101 pre-trained on ImageNet classification; and ResNet-50 hierarchical pretrained. We used Facebook Research's Detectron2 \cite{wu2019detectron2} library for training and evaluating object detection models. The randomly initialized backbone and backbones pretrained on ImageNet were readily available through the Detectron2 Model Zoo \cite{wu2019detectron2}.

For the hierarchical pretrained model, we used ResNet-50 backbone, fine-tuned from weights pretrained using the Simple Framework for Contrastive Learning of Visual Representations (SimCLR) \cite{DBLP:journals/corr/abs-2002-05709}. SimCLR is a contrastive learning framework that teaches a model to closely associate multiple augmented views of the same image. For our specific self-supervised learning approach, we implemented hierarchical pretraining, a framework that uses several stages of training, arranging each successive pretraining step to a dataset closely resembling the target labeled data \cite{reed2022self}. We trained our pretrained model using weights \cite{Downes_2023_WACV} that had been pretrained on Worldview-3 Maxar imagery, similar to our cell tower dataset (i.e., both are overhead object detection problems that use Maxar imagery). For the overall framework, we used Facebook Research's VIsion library for state-of-the-art Self-Supervised Learning (VISSL) \cite{goyal2021vissl} for model pretraining. 

\paragraph{Implementation Details.} \label{sec:implement}

For the training environment, we used an ml.g4dn.12x large instance on Amazon Web Services with four NVIDIA T4 GPUs. The batch size was set to 8 and trained for 12500 iterations for the pre-trained backbones and 90000 iterations for the randomly initialized backbone. Training took between 6 and 7 hours per experiment. For network optimization, we used the default Stochastic Gradient Descent optimizer with an initial learning rate between 0.02 and 0.2 with the best parameters (specified in Table \ref{tab:config} in the Appendix). We started the experiments with default configuration settings available within VISSL (hierarchical pretrained) or Detectron2 (random initialized or ImageNet pretrained). We then explored different normalization techniques, such as Frozen Batch Norm, Batch Norm, and Sync Batch Norm, Region of Interest (ROI) Heads, and frozen layers, to identify the configuration for optimal results. To augment the data, we used random horizontal flips to prevent model overfitting. Table \ref{tab:config} in the Appendix contains the full details of the training configurations used.
   

\section{Experiments and Results}

\paragraph{Baseline Results.}

To evaluate the performance of the ResNet backbones, we used the COCOEvaluator available from the Detectron2 library, which provides default evaluation metrics. We used Average Precision (AP)---traditionally called mean average precision or mAP---averaged across ten IoU thresholds. We also used AP at IoU=0.50 and 0.15 (AP@50 and AP@15, respectively) to compare performance of different model backbones. Further metric details can be found in the common objects in context (COCO) challenges documentation\footnote{\url{https://cocodataset.org/\#detection-eval}}. We evaluate using this lower IoU AP@15 because our primary aim is to locate cell towers within an image and we are not as concerned with determining each tower's exact bounding box.

We created train and test splits by performing an 80:20 split yielding 5023 images containing 6842 annotations for the training set and 1256 images containing 1713 annotations for the test set. The hierarchical pretrained model achieves an average precision AP@50 of 81.2 on the validation split of our dataset, with good performance across different geographies. Table \ref{tab:baseline}  lists the AP, AP@15, and AP@50 for our baseline hierarchical pretrained model and the backbones available from the Detectron2 Model Zoo. Figure \ref{fig:objdet} shows example predictions from the test set generated by the RestNet50-hierarchical pretrained model. These preliminary results, based on auto-generated bounding boxes, showed that our methods can detect most of the targets with a relatively low rate of false alarms. We observed that including negative chips in the Detectron2 model did not improve model performance.

\begin{figure} 
\begin{center}
   \includegraphics[width=0.9\linewidth]{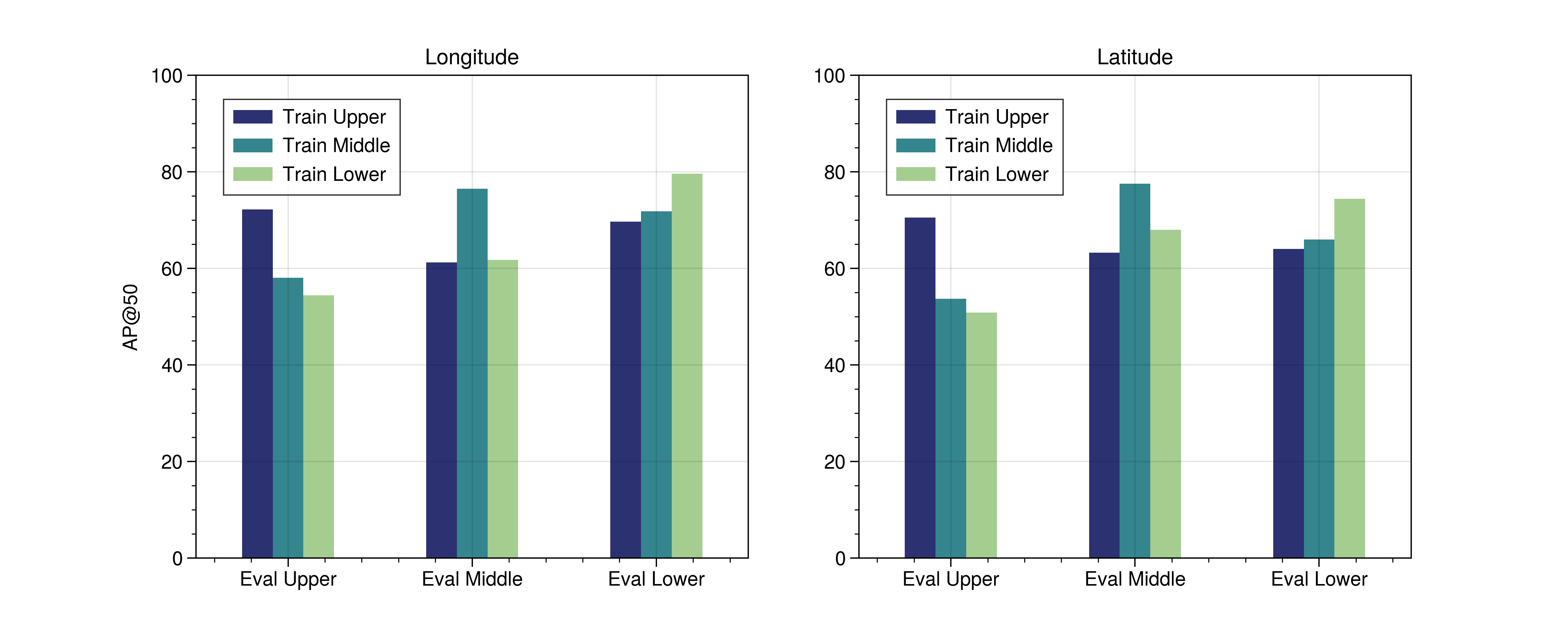}
\end{center}
   \caption{Out-of-sample and in-sample experimental results for longitude and latitude regions.}
\label{fig:oss}
\end{figure}

\paragraph{Region-Specific Models.}

We tested to see whether a region-specific model yielded performance similar to our baseline model. We stratified the annotations by latitude and longitude to define subsamples of annotations (shown in Table \ref{tab:regions_summary} in the Appendix). By partitioning the data with these specific latitudes and longitudes, the total number of annotations within the six partitions is roughly equal. We then trained the model on one region (e.g., lower latitude ranges) and tested it in regions for which it was not trained (e.g., middle and upper latitude ranges). 

We evaluated the baseline model (trained with data from all regions) on a subset of upper, middle, and lower latitude images. We found that the model performed best in the middle latitude region. There was improvement in performance between our baseline and region-specific results (shown in Table \ref{tab:regions}). We hypothesize that the increased performance of the baseline model is due to additional training data (roughly a factor of three times greater than data for the region-specific models), leading to better generalization on the test sets.


\begin{table}[t]
  \caption{AP@50 scores of baseline and region-specific models.}
  \label{tab:regions}
  \centering
  \begin{tabular}{ l ccc} 
    \toprule
    Evaluation & Baseline & Region-Specific (Latitude) & Region-Specific (Longitude)  \\
    \midrule
  Upper & 78.1 & 70.5 & 72.2 \\
   Middle & 84.8 & 77.5 & 76.5 \\ 
  Lower & 81.5 & 74.4 & 79.6 \\ 
    \bottomrule
  \end{tabular}
\end{table}

\paragraph{Out-of-Sample Results.}

We explored out-of-sample results to investigate whether the results are generalizable across latitude and longitude ranges. We trained a model on upper, middle, and lower latitude or longitude range annotations, then evaluated the model on these ranges. As expected, the in-sample results performed the best out of all of the results, as shown in Figure \ref{fig:oss}. The decrease in out-of-sample performance follows similar patterns between the longitude and latitude ranges. The out-of-sample results evaluated on upper and middle latitude and longitude annotations had the most marked decline in performance compared to the in-sample results. For the upper latitude and longitude images, this could be due to the backgrounds that include desert landscapes, such as in the Horn of Africa, that make the mast structure of cell towers difficult to see. In contrast, the smallest difference in out-of-sample and in-sample performance existed for the models evaluated on lower latitude and longitude samples. This was perhaps due to the lower latitude and longitude areas containing a mixture of backgrounds from upper and middle latitudes. 

\section{Impact and Next Steps}

The generalized object detector developed through this research is able to correctly identify cell phone towers across a vast geographic area. We proposed a semi-automated framework for developing a training dataset that reduces the time---and therefore cost--- associated with manual labelling. The cell tower object detector will help enable research on cell phone coverage in remote areas, which can guide improved placement of towers by mobile network operators. Partnerships between telecoms and non-profit organizations have already proven successful in installing new cell phone towers in rural areas and give local residents direct access to network coverage. Providing tower locations can help resolve cell coverage maps for last-mile populations and improve the delivery of a wide range of digital services. This work also helps enable cross-disciplinary research that uses cell tower location as a covariate for social and environmental modelling studies. In continued work, we plan to investigate performance of the detector across a broad area search and produce a dataset of cell tower locations across sub-Saharan Africa.

\section{Conclusion}

In this work, we presented a methodology to annotate overhead imagery using OSM that provides training data for an object detection model to identify cell towers in sub-Saharan Africa. Using Detectron2 and high-resolution imagery, we trained and tested a Faster R-CNN architecture on a dataset of roughly 6000 cell towers. We fine-tuned several ResNet model backbones from model weights that were pretrained using hierarchical pretraining. Our results indicate that bounding boxes set from a radius of 25 meters from the tower center point are adequate to produce good results that are generalizable in- and out-of-sample, and across different geographies. The results also suggest that using automated bounding boxes from center points can reduce the time and effort expended on labeling, without compromising model performance. Crowdsourced features in OSM can provide datasets for other objects of interest, accelerate model development, and reduce labeling burden.


\begin{ack}
This research was supported by the Department of Defense Science, Mathematics, and Research for Transformation (SMART) Scholarship SEED grant awarded to Dr. Natasha Krell. This document is approved for public release, NGA-U-2023-02104. All findings, opinions, conclusions and recommendations expressed in this article are those of the author(s) and do not necessarily reflect the views of the U.S. Government. 
\end{ack}

\medskip

{\small
\bibliographystyle{unsrtnat}
\bibliography{mybib}
}

\section*{Appendix A}

\subsection*{A.1 Chips Generation and Automated Image Annotation} \label{sec:auto}

Using Python and widely-used geospatial libraries, such as GeoPandas \cite{kelsey_jordahl_2020_3946761}, Rasterio \cite{rasterio}, and Shapely \cite{shapely2007}, we chipped our images and created axis-aligned bounding boxes from the centerpoints of the cell towers. We created non-overlapping chips from the Maxar images and OSM vector point coordinates. We set a point buffer of 25 meters to buffer features into polygonal axis-aligned bounding boxes, ensuring that the buffer radius was not more than one quarter of the chip size. We chipped each image to a 512 x 512 pixel JPEG. For each image ID, we selected one positive and one negative sample (i.e. chip). A total of 9769 chips were available for training comprising of 6279 positive chips with labels and 3400 negative chips.  

\setcounter{table}{0}
\renewcommand{\thetable}{A\arabic{table}}

\begin{table}[h]
  \caption{Number of chips and annotations in test and training positive image sets.}
  \label{tab:numbers}
  \centering
  \begin{tabular}{ l ccc} 
    \toprule
    Positive Samples & Test & Train & Total  \\
    \midrule
  Chips & 1256 & 5023 & 6279 \\
   Annotations & 1713  & 6842 & 8555 \\ 
    \bottomrule
  \end{tabular}
\end{table}

\begin{table}[h]
  \caption{Latitude and longitude ranges and sample size for region-specific models.}
  \label{tab:regions_summary}
  \centering
  \begin{tabular}{ l ccc} 
    \toprule
    Regions & Range & Annotations Train & Annotations Test  \\
    \midrule
  Upper Latitude & [-2\textdegree S, 14\textdegree N] & 2244 & 574 \\
   Middle Latitude & [-2\textdegree, -16.5\textdegree S] & 2334 & 510 \\ 
  Lower Latitude & [-16.5\textdegree, -28\textdegree S] & 2264 & 629 \\
   Upper Longitude & [18\textdegree, 31\textdegree E] & 2290 & 602 \\
   Middle Longitude & [31\textdegree, 41\textdegree E] & 2232 & 520 \\
   Lower Longitude & [41\textdegree, 58\textdegree E] & 2320 & 591 \\
    \bottomrule
  \end{tabular}
\end{table}

\begin{landscape}
\begin{table}
  \caption{Configuration of training parameters for Faster RCNN model training in Detectron2. Prior to training, the setup requires a configuration for the specific backbone. We generally start with the configuration setup available in Detectron2 then iterate on select parameters to improve performance. RN-50 and RN-101 stands for ResNet-50 and ResNet-101, respectively; HPT stands for hierarchical pretrained; RI stands for random initialize; INT stands for ImageNet pretrained.}
  \label{tab:config}
  \centering
  \begin{tabular}{ l cccc} 
    \toprule
    Model Parameter & RN-50-HPT & RN-50-RI & RN-50-INT & RN-101-INT \\
    \midrule
  MODEL.WEIGHTS & HPT & - & INT RN-50 & INT RN-101  \\
   MODEL.RESNETS.DEPTH & 50 & 50 & 50 & 101 \\ 
  MODEL.RESNETS.NORM & SyncBN & SyncBN & FrozenBN & FrozenBN\\
  MODEL.ROI\_HEADS.NAME & Res5ROIHeadsExtraNorm & Res5ROIHeads & Res5ROIHeads & Res5ROIHeads \\
  MODEL.BACKBONE.FREEZE\_AT & 0 & 0 & 2 & 2 \\
  SOLVER.BASE\_LR & 0.15 & 0.02 & 0.02 & 0.02 \\
  SOLVER.STEPS & 9500 & (60000, 80000) & 9500 & 9500 \\
  SOLVER.MAX\_ITER & 12500 & 90000 & 12500 & 12500 \\
   
    \bottomrule
  \end{tabular}
\end{table}
\end{landscape}


\end{document}